# BigCilin: An Automatic Chinese Open-domain Knowledge Graph with Fine-grained Hypernym-Hyponym Relations

Ming LIU[12], Yaojia LV[1], Jingrun ZHANG[1], Tianwen JIANG[1], Ruiji FU[3], Bing QIN[*12]

[1] Harbin Institute of Technology, Harbin, China
[2] PENG CHENG Laboratory, Shenzhen, China
[3] Kuaishou Technology. China
mliu@ir.hit.edu.cn, yjlv@ir.hit.edu.cn, jrzhang@ir.hit.edu.cn,
twjiang@ir.hit.edu.cn, furuiji@kuaishou.com, bqin@ir.hit.edu.cn

## Abstract

This paper presents BigCilin, the first Chinese open-domain knowledge graph with fine-grained hypernym-hyponym relations which are extracted automatically from multiple sources for Chinese named entities. With the fine-grained hypernym-hyponym relations, BigCilin owns flexible semantic hierarchical structure. Since the hypernym-hyponym paths are automatically generated and one entity may have several senses, we provide a path disambiguation solution to map a hypernym-hyponym path of one entity to its one sense on the condition that the path and the sense express the same meaning. In order to conveniently access our BigCilin Knowledge graph, we provide web interface in two ways. One is that it supports querying any Chinese named entity and browsing the extracted hypernym-hyponym paths surrounding the query entity. The other is that it gives a top-down browsing view to illustrate the overall hierarchical structure of our BigCilin knowledge graph over some sampled entities.

## 1 Introduction

Open Information extraction (Open-IE) is an important task in natural language processing (NLP) field. Various Open-IE systems such as OLLIE (Mausam et al., 2012), NELL (Carlson et al., 2010; Mitchell et al., 2018) and PATTY (Nakashole et al., 2012) are provided and widely applied to acquire (*entity*, *relation*, *entity*) triples. As connecting the triples via overlapping entities, a graph with entities as nodes and relations as arcs can be formed. This graph is nominated as Knowledge Graph, and applied in many NLP related tasks such Q&A, Dialogue, etc.

Recent methods for knowledge graph construction lay on entity level, which means they extract open-domain triples automatically, but leave the task of schema construction (i.e. the extraction of hypernyms for entities and the construction of hypernym-hyponym paths over the hypernyms) via manual operation (Wu et al., 2020). Schema is the backbone of knowledge graph and controls the types of the entities involved by knowledge graph. Semantic hierarchy (i.e. schema) construction for knowledge graph has been studied by many researchers. Most of researches are conducted based on manually built semantic resources such as WordNet (Miller, 1995). Such kind of schema has good structure and high accuracy, but its coverage is limited and only includes some coarse-grained concepts. Taking one famous Knowledge graph YAGO (Suchanek et al., 2008) for example, its schema is constructed based on the type categories in Wikipedia, while these types are just learned from WordNet. This situation causes the limitation of the types of entities involved by YAGO. As we known, DBpedia is almost the largest open-source knowledge graph, whereas the coverage of its schema is still limited by the scope of Wikipedia.

To this end, we propose BigCilin, the first Chinese open-domain knowledge graph, whose schema is automatically generated. Different from traditional methods which construct knowledge graph from a top-down angle, that indicates their schemas are formed manually and the entities and relations are collected following the predefined schema (Sarhan et al., 2021), our BigCilin is formed from a bottom-up angle. We collect as many entities and their relations as possible. The schema of BigCilin is automatically constructed based on the acquired entities. Section 2.1 and 2.2 give the detailed process.

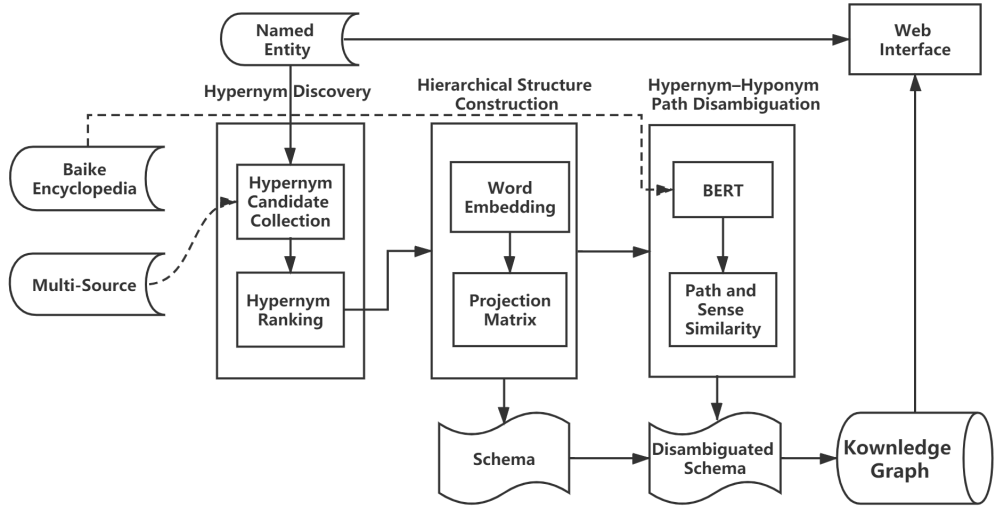

Figure 1. The overview architecture of our BigCilin. (Given a named entity, we collect the hypernym candidates from multiple sources and acquire the meaningful and valuable hypernyms using a ranking algorithm. Then, the approach of hierarchical structure construction aims to automatically detect hypernym-hyponym relations among the detected hypernyms via learning a projection matrix based on word embedding. In the end, the approach of hypernym-hyponym path disambiguation aims to map the hypernym-hyponym paths to one sense of the entity on the condition that they express the same meaning.)

One Chinese named entity may own several senses. For example, "apple" just indicates a fruit, a mobile phone, or a cooperation. The schema of BigCilin is constructed automatically all by input named entities, while these input entities are not disambiguated. Given one input entity "苹果(apple)", our hypernym-hyponym extraction process will build the hypernym-hyponym paths indicating all the meanings of "苹果(apple)". To facilitate the use, we collect the senses for each entity included by our BigCilin. For each entity, we designate the hypernym-hyponym paths to its one sense when they indicate the same meaning. The details can be found in Section 2.3.

Our BigCilin can be accessed through URL "http://bigcilin.com". We provide web interface in two ways. One way is query type. Users can input one query word (not limited to entity). Our BigCilin will output the automatically generated hypernym-hyponym paths and the detailed (*entity*, *relation*, *entity*) tripes for this query word. The other way is browser type. Users can browse the overall schema of BigCilin.

Our BigCilin knowledge graph has two innovations: (1) it is the first Chinese knowledge graph with automatically constructed schema of open-domain fine-grained hypernym-hyponym relations and (2) it provides an automatic disambiguation solution via designating the hypernym-hyponym paths to one sense of the entity when they indicate the same meaning.

## 2 System Description

Figure 1 shows the overview architecture of our BigCilin knowledge graph (or can be called system). The entities are collected from multiple sources, including NER algorithm conducted on web texts, dictionary, and online encyclopedia. They are treated as input. In order to automatically construct the schema with fine-grained hypernym-hyponym relations, we can roughly divide the construction process into three steps: Hypernym Discovery, Hierarchical Structure construction, and Hypernym-Hyponym Path Disambiguation.

### 2.1 Hypernym Discovery

The aim of Hypernym Discovery is to acquire the fine-grained hypernyms for one given entity. Our approach is composed of two major steps: hypernym candidate collection and hypernym ranking, as reported in (Fu et al., 2013). In the first step, we collect hypernym candidates from multiple sources. Given a named entity, we search it in a searching engine and extract nouns (or noun phrases) with high concurrence frequency as the hypernym candidates. In detail, we count the concurrence frequency between the target entity and other words in the returned snippets, and choose the top

$N$ ($N$ is set to 10) nouns (or noun phrases) as the hypernym candidates. We also collect the category tags of the entity from Baike encyclopedia[1] as the hypernym candidates. The manually annotated encyclopedia category tags are the important clues to identify entity's meaning. In addition, the head words of entities are often the hypernyms of entities according to the morphology of Chinese word. For example, the head word of "皇帝企鹅 (Emperor Penguin)" is "企鹅 (Penguin)", which indicates that this entity is a kind of penguins. Thus, we put the head word of the given entity into the candidates. We combine all these hypernym candidates together as the input of the second step.

In the second step, we filter incorrect hypernyms out of the candidate set. We view this task as a ranking problem and build a statistical ranking model by assembling the results coming from three ranking algorithms. They are Support Vector Machine (SVM) with a linear kernel, SVM with a radial basis function (RBF) kernel, and Logistic Regression (LR). We also propose a heuristic strategy to collect training data to train each ranking algorithm. The details can be found in (Fu et al., 2013). The coverage rate of our approach reaches 93.24% out of all the hypernyms for the sampled entities.

## 2.2 Hierarchical Structure Construction

The aim of Hierarchical Structure Construction is to build the schema of our BigCilin knowledge graph automatically. From the previous step, we find many fine-grained hypernyms. The key of this step is to detect as many hypernym-hyponym relations as possible among the detected hypernyms. We provide a solution based on word embedding, as reported in (Fu et al., 2014), which learns a projection matrix to identify whether an input word pair has hypernym-hyponym relation or not. Intuitively, given one word $x$ and its one hypernym $y$, there exists a projection matrix $\Phi$ so that $y = \Phi x$. However, as entities have types, the hypernym-hyponym relations for different entity types should be modeled by different projection matrixes. For example, the matrix used to model the hypernym-hyponym relation for fruit is distinguished from the one for animal. For this reason, given $x$ and $y$, we first cluster $y - x$ and denote their cluster as $C_k$. Then, we apply equation (1) to obtain the corresponding projection matrix $\Phi_k$ for $C_k$.

---

[1] https://baike.baidu.com/

$$\Phi_k^* = \underset{\Phi_k}{\text{argmin}} \frac{1}{N_k} \sum_{(x,y) \in C_k} \|\Phi_k x - y\|^2 \quad (1)$$

where $N_k$ is the amount of input word pairs in the $k^{th}$ cluster $C_k$.

Given two words $x$ and $y$, if $y$ can be considered as a hypernym of $x$, only the projection matrix $\Phi_k$ puts $\Phi_k x$ close enough to y. Besides, if a relation circle appears, we remove or reverse the link between the weakest word pair to make the structure of hypernym-hyponym relations as a tree. In detail, if a circle has only two nodes, we remove the weakest link. If a circle has more than two nodes, we reverse the weakest link to indicate the indirect hypernym-hyponym relation expressed by the remaining hypernym-hyponym path.

## 2.3 Hypernym-Hyponym Path Disambiguation

Previous approach can construct the fine-grained hypernym-hyponym paths for input entity. However, since the input entity is not disambiguated, the constructed hypernym-hyponym paths indicate all the meanings of input entity. Figure 2 just shows an example. The input entity is "苹果(apple)", and the constructed hypernym-hyponym paths indicate all the meanings of "苹果(apple)". Our disambiguation aims to separate the hypernym-hyponym paths according to the senses of the entity. As Figure 2 shows, we designate two paths "苹果(apple)→水果(fruit)→食品(food)→物(thing)" and "苹果(apple)→水果(fruit)→植物(Plant)→生物(biology)→物(thing)" to the sense "蔷薇科苹果署果实(rosaceae fruit of apple type)" of "苹果(apple)". We first collect the senses of entities from Baike encyclopedia. Then, we treat one sense of the entity and one hypernym-hyponym path as two strings. The string for the sense is composed by several phrases lined in the alphabetical order. The phrases are the relation or attribute triples without the head entity (the relation and attribute triples are also extracted automatically, whereas, due to the limited length we omit the details). One example phrase for the sense "蔷薇科苹果署果实(rosaceae fruit of apple type)" of the entity "苹果(apple)" is "性味(property and flavor)/微甜(slightly sweet)", which can be found in Figure 2 surrounded by blue box.

We apply pretrained BERT (Devlin et al., 2019) to encode the two strings to get their representations, and utilize Cosine to score the similarity between two strings. We annotate 6,000 positive samples and 6,000 negative samples as training data. The disambiguated results are acceptable and can be viewed through BigCilin website.

## 3 Demonstration

A demo system is provided. It contains two access modules: Query module and Browse module.

### 3.1 Query module

Query module supports querying any Chinese named entity. One example is shown in Figure 2. The senses of the input entity and the hypernym-hyponym hierarchical structure for this entity are illustrated. When clicking one sense, the disambiguated hypernym-hyponym paths corresponding to this sense are highlighted. Besides, the relation and attribute triples related to the clicked sense are also listed. If the input entity is included by our BigCilin, the previous data are retrieved. If not, the data are automatically generated or extracted from multiple sources as told in Section 2.

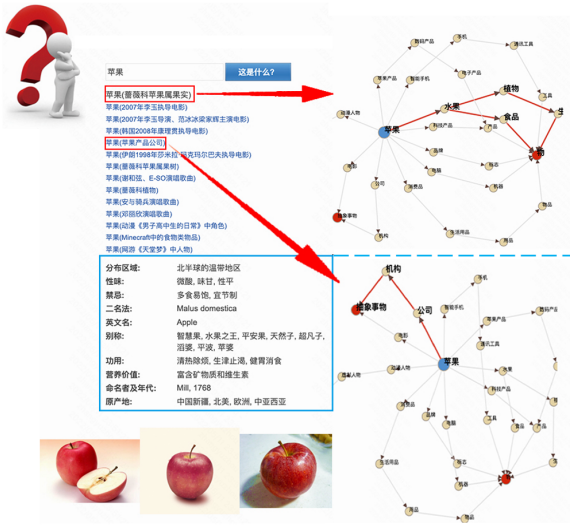

Figure 2. An example of query module on the query "苹果(apple)". (When input query entity, the senses of the entity and the relation triples belonging to this sense are returned. Besides, the hypernym-hyponym hierarchical structure of the entity is formed with the disambiguated paths highlighting.)

### 3.2 Browse module

Browse module supports users to browse the overall schema of our BigCilin. One example is shown in Figure 3. Till now, our BigCilin has over 30 million entities and 180 thousand hypernyms. The schema of our BigCilin is complex and large-scale. Thus, to avoid memory overflow, we only demonstrate a part of the schema over some sampled entities. In addition, we have shared the core data of our BigCilin in OpenKG[2] including common entities and their fine-grained hypernyms.

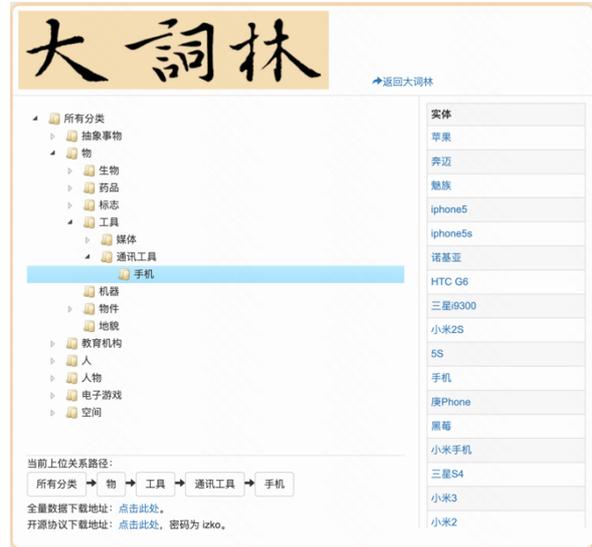

Figure 3. An example of browse module. (The whole schema of BigCilin is shown on the left part, and the right part gives the detailed entities related to the clicked hypernym-hyponym path)

## 4 Conclusion

In this paper, we introduce BigCilin, the first automatically formed Chinese open-domain knowledge graph. The schema of BigCilin is constructed automatically, and contains fine-grained hypernym-hyponym relations, which can support many downstream NLP tasks. Besides, we also provide a path disambiguation solution to map a hypernym-hyponym path of one entity to its one sense on the condition that the path and the sense express the same meaning. In order to provide user friendly access, we provide two kinds of web interfaces. One can be accessed by inputting a query word and then browsing the disambiguated hypernym-hyponym paths. The other can be used to demonstrate the schema of our BigCilin from a top-down view.

---

[2] http://www.openkg.cn/dataset/hit